\documentclass[letterpaper,10pt,conference]{ieeeconf}
\IEEEoverridecommandlockouts
\usepackage{amsmath,amssymb,amsfonts}
\usepackage{algorithmic}
\usepackage{graphicx,graphics}
\usepackage{caption,subcaption}
\usepackage{xcolor}
\usepackage{url}
\usepackage{booktabs}

\title{\LARGE \bf
Learning to Predict Grip Quality from Simulation: \\ Establishing a Digital Twin to Generate Simulated Data for a Grip Stability Metric
}

\author{Stefanie Wucherer$^{1}$, Robert McMurray$^{2}$, Kok Yew Ng$^{2}$, Florian Kerber$^{1}$
\thanks{$^{1}$S. Wucherer and F. Kerber are with the Technology Transfer Center of the University of Applied Sciences Augsburg, Germany.
        {\tt\small \{stefanie.wucherer, florian.kerber\}@hs-augsburg.de}.}%
\thanks{$^{2}$K. Y. Ng and R. McMurray are with the School of Engineering of Ulster University, UK.
        {\tt\small \{mark.ng, rj.mcmurray\}@ulster.ac.uk}.}%
}

\begin{document}

\onecolumn{\textcopyright 2023 Ulster University/Technology Transfer Center (University of Applied Sciences Augsburg). Personal use of this material is permitted. Permission from the authors must be obtained
for all other uses, in any current or future media, including reprinting/republishing this material for
advertising or promotional purposes, creating new collective works, for resale or redistribution to
servers or lists, or reuse of any copyrighted component of this work in other works.}
\newpage
\twocolumn
\maketitle
\thispagestyle{empty}
\pagestyle{empty}

\begin{abstract}
A robust grip is key to successful manipulation and joining of work pieces involved in any industrial assembly process. Stability of a grip depends on geometric and physical properties of the object as well as the gripper itself. Current state-of-the-art algorithms can usually predict if a grip would fail. However, they are not able to predict the force at which the gripped object starts to slip, which is critical as the object might be subjected to external forces, e.g. when joining it with another object. This research project aims to develop a AI-based approach for a grip metric based on tactile sensor data capturing the physical interactions between gripper and object. Thus, the maximum force that can be applied to the object before it begins to slip should be predicted before manipulating the object. The RGB image of the contact surface between the object and gripper jaws obtained from GelSight tactile sensors during the initial phase of the grip should serve as a training input for the grip metric. To generate such a data set, a pull experiment is designed using a UR 5 robot. Performing these experiments in real life to populate the data set is time consuming since different object classes, geometries, material properties and surface textures need to be considered to enhance the robustness of the prediction algorithm. Hence, a simulation model of the experimental setup has been developed to both speed up and automate the data generation process. In this paper, the design of this digital twin and the accuracy of the synthetic data are presented. State-of-the-art image comparison algorithms show that the simulated RGB images of the contact surface match the experimental data. In addition, the maximum pull forces can be reproduced for different object classes and grip scenarios. As a result, the synthetically generated data can be further used to train the neural grip metric network.
\end{abstract}


\section{INTRODUCTION}
In robotic handling, joining, and assembly processes such as those used in the gearbox assembling industry, more and more flexibility is demanded of the overall robotic system in terms of performance, reliability, and robustness, which is in line with the aims of Industry 4.0. \cite{tantawi2019i40, HU2011prodvar} For new solutions to be integrated into industrial production lines, secure handling of objects is of crucial importance. As a result, the evaluation of grip stability is a first step to ensure flexible and secure automated assembly processes. The stability of the grip itself is governed by several physical properties, which include the geometry and texture of the object, as well as properties of the robot system such as the geometry of the gripper or the force of which the object is gripped. Therefore, to decide whether or not a grip is stable, these properties have to be measured. To this end, several different types of tactile sensors have been developed to perform these measurements. \cite{yuan2017gelsightbasic,rosle2020hall,neto2021} In this work, the optical GelSight R1.5 sensor \cite{yuan2017gelsightbasic} is used which measures the contact surface topology thus capturing the relevant properties for grip analysis.

The R1.5 has been applied successfully by other research groups to predict if a grip is stable to pick up an item, or if it will slip out when being picked up. \cite{calandra2018gelsight} During assembly, however, additional forces are applied to the object, e.g. due to contact between peg and hole in joining processes. Therefore, it is not enough to predict the success of manipulation, but also how much force can be applied to the object during the grip. To predict this force, which is essentially the static friction force between object and gripper jaws, this project aims to utilize a neural network rather than to physically model the highly non-linear friction characteristics. To make the network as robust as possible against variations in object geometry and texture, the training sample set needs to cover as wide a variety of grip configurations as possible. To automatically generate such a data set a pull experiment has been designed in which a spatially fixed object is gripped by a robot and the tactile sensor images of the initial grip are stored. The robot begins pulling on the object in a certain direction, increasing the force until the jaws slip off, generating a label that can be used as reference for training the neural grip metric network. 

Using this real life experiment to generate training data, however, is time consuming. For this reason, a digital model of the experimental setup has been generated in a virtual environment, with the intention to upgrade this to a digital twin in the future.\cite{melesse2020digital} In this paper, the structure and parameterization of the simulation will be presented. The goal is to reproduce both the surface images and maximum pull forces which will serve as labeled inputs for the neural grip metric network. The physics simulation framework PyBullet \cite{coumans2021PyBullet} was chosen since it can model the physical properties of all the mechanical components, i.e. the robot and the gripping objects. The properties of the tactile sensor were modelled using the Taxim\_robot \cite{si2022taxim_robot} library of the Taxim \cite{si2022taxim} software package. The experiment was implemented as a custom environment of the OpenAI gym framework. This allows for quick interchange of robot steering policies, including the use of neural networks, if needed. As an industrial use case, parts of a planetary gearbox were selected to compare simulated and measured data and to evaluate the accuracy of the digital model. In a first step presented here, to make the comparison more straightforward and less resource intensive, only pulls in the positive z-direction are considered.

This paper is organised as follows: Section \ref{RelatedWork} introduces related work found in the literature. Section \ref{sec:experiment} describes the experiment used to generate the training data for the grip metric. Section \ref{sec:digital-model} presents the design of the digital model and the adjustments necessary to generate useful data. Then, Section \ref{sec:comparison} compares simulated data set against its actual counterpart for a selection of different objects. Finally, Section \ref{sec:outlook} provides some conclusions and the outlook for the future usage of the proposed digital twin.

\section{RELATED WORK}\label{RelatedWork}
Previously, tactile sensors have been used to analyse grip stability. Both the simulation models for such systems and derived metrics will be summarized in the following.

\subsection[Simulation of tactile sensors]{Simulation of Vision-Based Tactile Sensors }\label{sec:lit-simulation}
Simulations of high-resolution outputs of vision-based tactile sensors such as the GelSight, Digit, or Omnitact were not possible in the past years since common robotic simulators implemented a rigid body model, which made the modelling of contact of an elastic body not possible. TACTO \cite{si2022taxim} implemented a model where the normal force obtained from the physics simulator PyBullet is used to adjust the penetration depth of the object in the pyRender scene, which is then used to obtain an RGB image. Agarwal \textit{et al.} \cite{agarwal2021}  designed a similar approach using a different renderer where the optical properties such as depth and RGB image of the contact volume are determined via pyRender and by adjusting sensor parameters. In the approach of Taxim by Si \textit{et al.} \cite{si2022taxim}, a lookup table is generated via a calibration procedure, which establishes a positional relationship between depth gradient and pixel intensity. For exemplary depth images with clearly defined positions, forces, and geometries, above-average results could be generated in comparison between reality and simulation. Calandra \textit{et al.} \cite{calandra2018gelsight} uses the same principle in the Gazebo simulator. However, Taxim is the only simulator capable of simulating marker motion fields via the linear displacement and superposition principle.

\subsection{Stability Determined by Tactile Sensing}\label{sec:lit-tactile-sensing}
Hogan \textit{et al.} \cite{hogan2018tactileRegrasp} developed an approach for grip stability based on optical tactile sensors where a robot grips a part and shakes it after the grip. The resulting images of the tactile sensor are tagged with a label resulting from the experiment. The entire data set was trained using a ResNet 50 network. Then, a regrasping strategy was introduced by Calandra \textit{et al.} \cite{calandra2018gelsight} where a deep multimodal neural network was trained to predict grasp adjustments on an initially executed grasp. With a pressure sensor with a resolution of $4 \times 7$ pixels, Schill \textit{et al.} \cite{schill2012} trained a support vector machine (SVM) for data acquired via continuously measuring the grasp output during the grasping process. Si \textit{et al.} \cite{si2022taxim_robot} used simulated data to train a neural network to predict successful grasp execution and tested the performance of the trained network in a real world setup. The label used for training was binary (success/fail) and topological information was only available for one side of the object, as only one sensor was used to obtain tactile information. Studying the effect of both continuous, force-based labels and two-sided information on the stability prediction is one of the aims of this work.

\section[EXPERIMENT FOR TRAINING DATA GENERATION]{EXPERIMENTAL SETUP}\label{sec:experiment}
To be able to generate a reliable data set, a fully automated pull experiment is devised. It records an image of the contact surface between object and gripper and traces the pull force until slippage is detected. 

\subsection{Robotic Setup}\label{sec:robot_setup}
\begin{figure}[t!]
\centering
{\includegraphics[width=\linewidth]{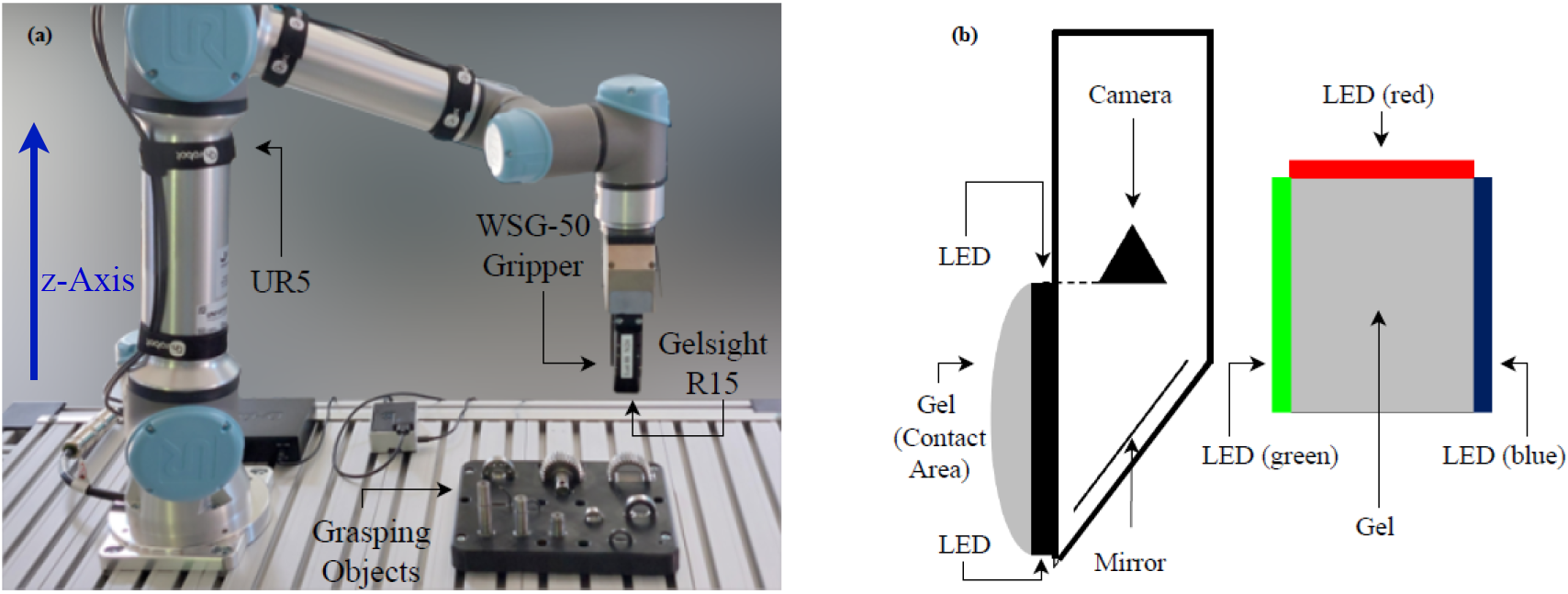}}
\caption{(a) The actual demonstrator system. (b) The schematic view of the GelSight sensor.}\label{fig:setup}
\end{figure}

The mechanical setup of the measuring system consists of a UR5e robot, the WSG 50 parallel gripper, and the GelSight's vision-based tactile sensors R1.5  (see Figure \ref{fig:setup}a). The UR5e is controlled using the Universal Robot Real Time Data Exchange interface, while the WSG 50 parallel gripper is controlled via the built-in telnet interface. The sensors measure the topology of the grip by taking a picture of the surface of a highly elastic gel, which is illuminated from three different directions using three different colour channels. To make the sensor as compact as possible, the camera is situated inside the R1.5 fingers pointing towards the finger tip, where a mirror is placed such that the image of the gel is reflected towards the camera. The image is then cropped and the effect of the mirror is algorithmically removed, thus generating as output of the sensor an RGB image with a resolution of $320 \times 240$ pixels. Figure \ref{fig:setup}b shows a schematic view of the sensors.

\subsection{Experiment Procedure}\label{sec:experimental_procedure}
\begin{figure}[t!]
\centering
\includegraphics[width=0.98\linewidth]{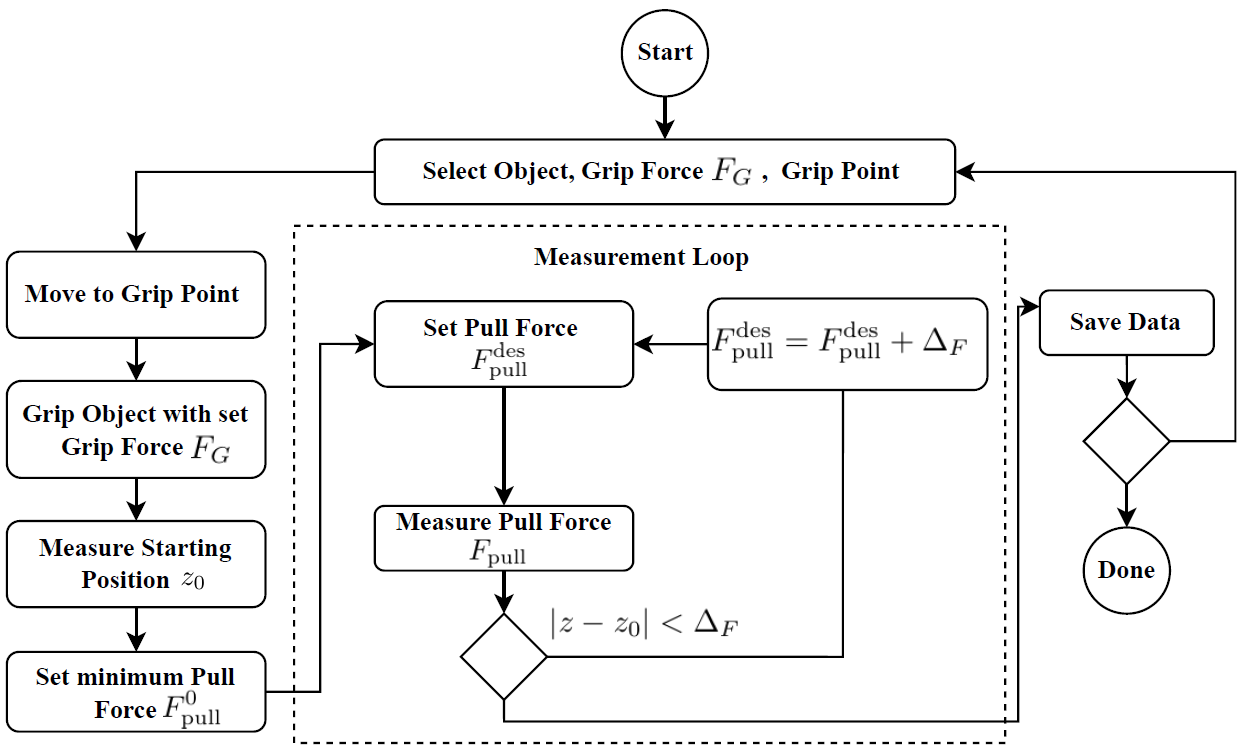}
\caption{Flowchart for the data-taking procedure.}\label{fig:flowchart_experiment}
\end{figure}

 The procedure of the experiment is as follows:
\begin{enumerate}
    \item A robot grips a rigid object which is fixed in all directions and orientations on a mounting plate. 
    \item Grip images on both gripper jaws are recorded and stored.
    \item The robot starts to pull upwards along the z-axis with a step-like pull force $F_{\rm pull}^{\rm des}(t)$
    \begin{equation*}
         F_{\rm pull}^{\rm des}(t)=\left\{\begin{array}{lr}F_{\rm pull}^0\,&t\in[0,\Delta_t)\\
         F_{\rm pull}^0+i \Delta_F,&t\in[i\Delta_t,(i+1)\Delta_t)\end{array}\right.
    \end{equation*}
    with constant force increments of $\Delta_F$ increased every $\Delta_t$~seconds.
    \item  The actual pull force $F_{\rm pull}(t)$ is measured and logged together with the target force $F_{\rm pull}^{\rm des}(t)$. 
    \item As soon as the displacement of the robot in z-direction exceeds the threshold $\Delta z$, the experiment is terminated.
\end{enumerate}
The steps 1--5 above are repeated for different classes of objects, different grip forces $F_G\in [5,80]$~N and different grip points on the objects. Multiple readings are taken for the same configuration to eliminate statistical fluctuations in the measurements. The flowchart of the experimental procedure is shown in Figure \ref{fig:flowchart_experiment}.

\subsection[Coupling of tactile info and stability]{Coupling Tactile Measurements and Pull Forces}\label{sec:label_calc}
The aim of the experiment is to correlate the sensor images of the contact surface with the maximum pull force $F_{\rm pull}^{\rm max}$ that can be applied to the object without slipping. The resulting metric should produce reliable predictions for rigid  objects independent of their material, texture and friction properties.

\begin{figure}[t!]
\centering
\begin{subfigure}[b]{0.475\columnwidth}
    \centering
    \includegraphics[width=\columnwidth]{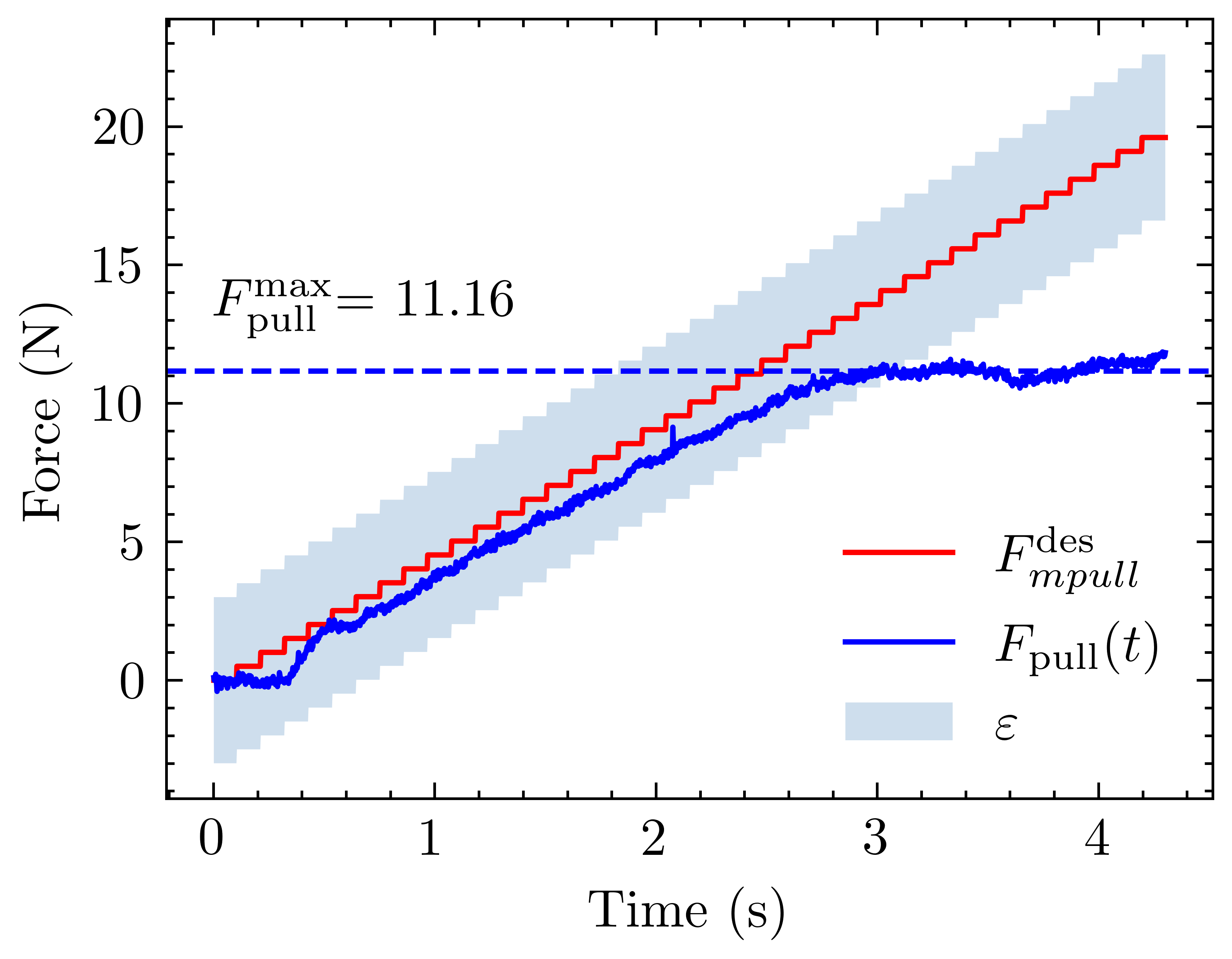}
    \caption{ }
\end{subfigure}
\hfill
\begin{subfigure}[b]{0.475\columnwidth}
    \centering
    \includegraphics[width=\columnwidth]{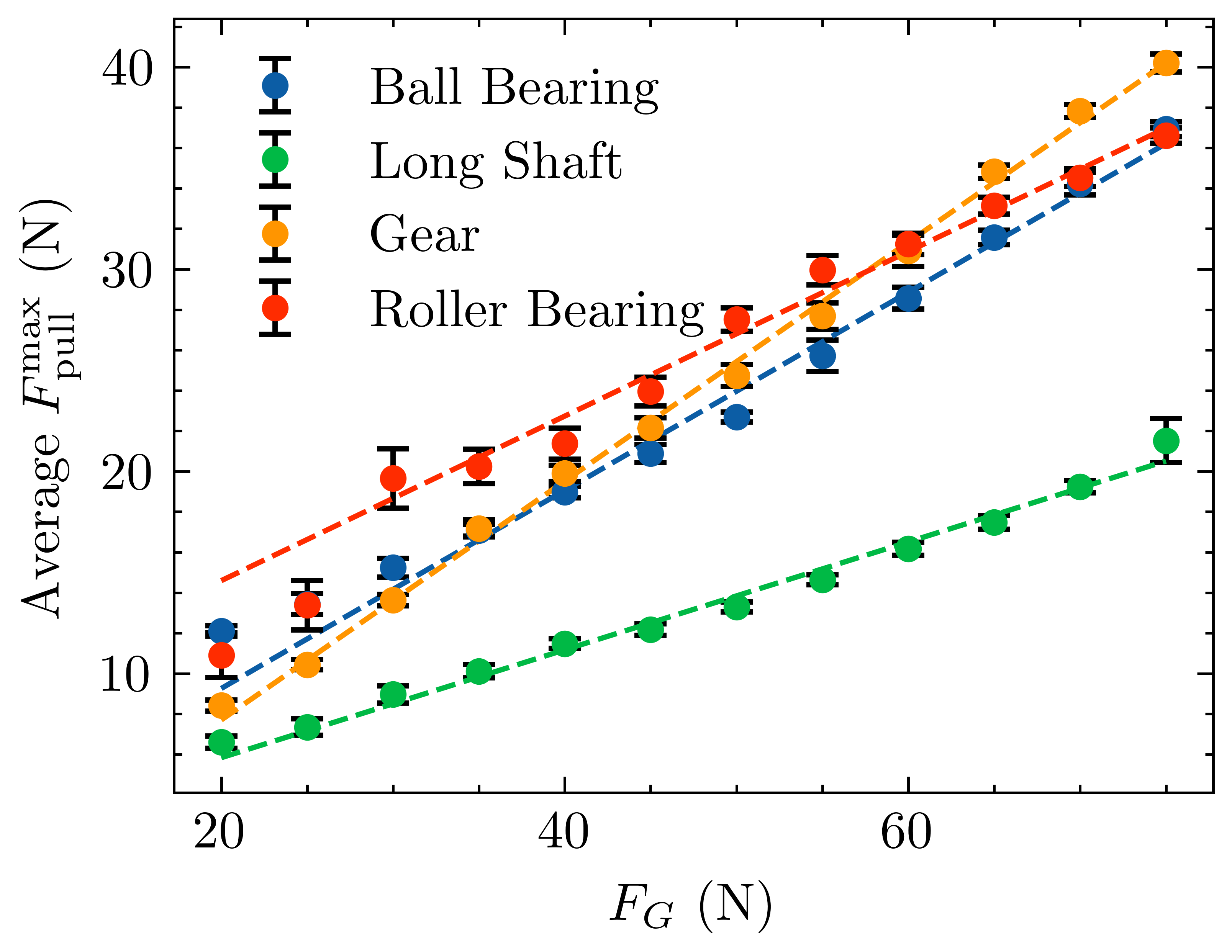}
    \caption{ }
\end{subfigure}
\caption{(a) The pull force profile of the grasped long shaft with a grip force of $F_G=40$~N.  (b) The averages of the maximum pull force for four different parts with different grip forces. Error bars are RMS.}
\label{fig:max_force_calculation}
\end{figure}

First,  the maximum force $F_{\rm pull}^{\rm max}$ before the robot starts to slip is determined from the experiment performed in Section \ref{sec:experimental_procedure}. Figure \ref{fig:max_force_calculation}a shows the plot of a typical experimental configuration, where the target pull force $F_{\rm pull}^{\rm des}(t)$ is plotted compared with the actual measured force $F_{\rm pull}(t)$ of the force-torque sensor in the last joint of the robot in the z-direction. The desired force and the actual force are offset since the drives of the robot have a reaction time before they can reach the set force. 
Using the threshold $\varepsilon = 3$~N, the
maximum pull force $F_{\rm pull}^{\rm max}$ up to which the robot can still hold the object can thus be computed as
\begin{eqnarray}
T_{\rm slip} &=& \arg \max_t\left\{\left|{F}_{\rm pull}(t)-F_{\rm pull}^{\rm des}\right|<\varepsilon\right\} \\
F_{\rm pull}^{\rm max}&=&F_{\rm pull}(T_{\rm slip})\,.
\end{eqnarray}

To determine the stability of the measurement per grip configuration (grip force, object, grip position and orientation), $10$ repetitions are performed on the validation data set using objects shown in Table \ref{tab:objects}. The mean and root mean square (RMS) are then calculated (see Figure \ref{fig:max_force_calculation}b. The influence of the individual grip configuration is clearly visible: the shaft gripped on the lateral surface exhibits much smaller friction than objects that are gripped on the sides such as gears and ball bearings. Hence, the maximum pull forces of the shaft are up to $20$~N lower compared to the other objects.
\begin{table}[t!]
\caption{Images of parts used for grip experiment (top) and their corresponding CAD (bottom) used in the simulation. }\label{tab:objects}
\centering
\begin{tabular}{cccc} 
 \toprule
 \textbf{Roller Bearing} & \textbf{Gear} & \textbf{Long Shaft} & \textbf{Ball Bearing}\\
 \midrule
\includegraphics[width=0.18\linewidth]{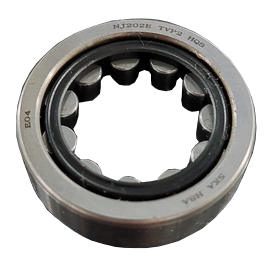}&\includegraphics[width=0.18\linewidth]{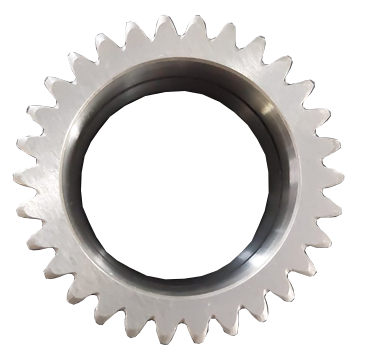}&\includegraphics[width=0.18\linewidth]{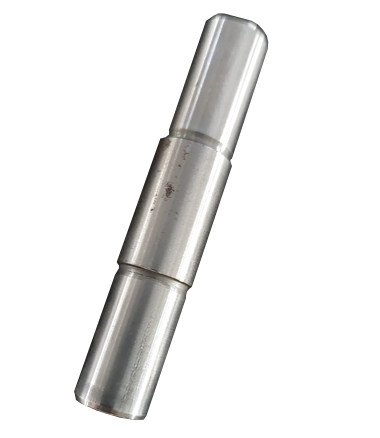}&\includegraphics[width=0.18\linewidth]{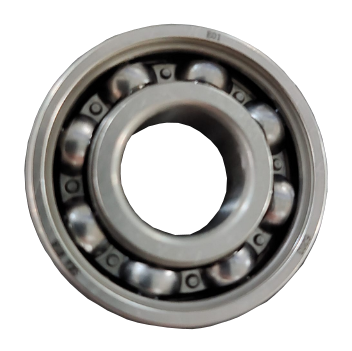}\\
\midrule
 \includegraphics[width=0.18\linewidth]{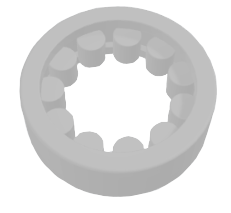}&\includegraphics[width=0.18\linewidth]{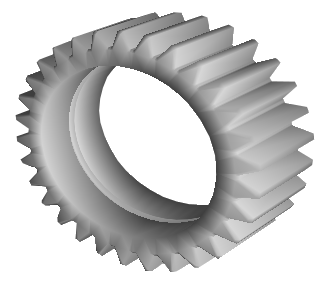}&\includegraphics[width=0.18\linewidth]{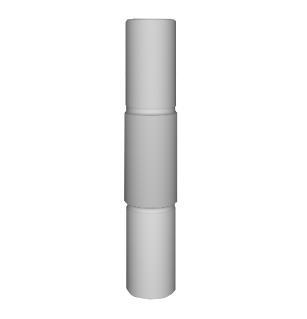}&\includegraphics[width=0.18\linewidth]{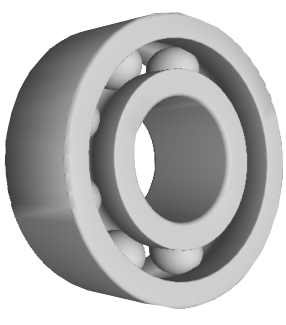}\\
\bottomrule
\end{tabular}
\end{table}

\section{DESIGNING THE DIGITAL MODEL}\label{sec:digital-model}
\begin{figure}[t!]
\centering
\includegraphics[width=.95\linewidth]{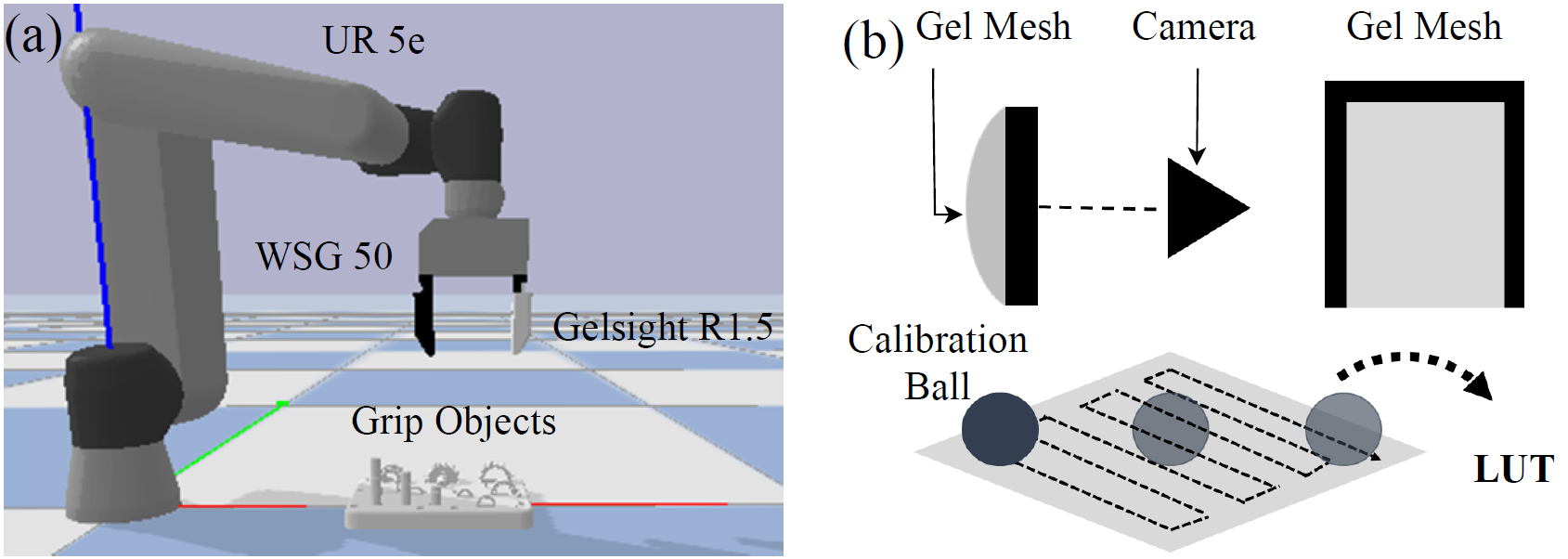}
\caption{(a) The environment in the digital model. (b) Schematic overview of the simulated sensor (note the absence of mirror and lights) and the calibration procedure scheme.}\label{fig:sim_env}
\end{figure}

To implement the digital model, the simulation has to be parameterized correctly. 
Figure \ref{fig:optical_sim} depicts the interaction of different simulation frameworks to obtain the multi-physics model of the experimental setup. A screenshot can be seen in Figure \ref{fig:sim_env}a.
In the following, the setup of the different physical domains is explained in detail.

\subsection{Perceptual Modelling}\label{sec:perception_modelling}
The GelSight R1.5 sensor had to be modeled in Taxim. The perception model consists of 
\begin{itemize}
    \item the geometry of the gel as \texttt{.stl} mesh,
    \item the pose of the camera in 3D (extrinsic parameters),
    \item the focal lengths of the camera (intrinsic parameters),
    \item the field of view of the camera and
    \item the resolution of the image $320\times 240$ pixels.
      \end{itemize}
Thus, a simplified optical path without the tilted mirror was modeled, see Figure \ref{fig:sim_env}b).

To incorporate the depth information, a calibration step had to be performed with a ball of diameter $3.94$~mm. The contact surface of this calibration ball was taken from various angles and positions. Thus, a lookup table (LUT) could be generated to simulate the color image from the known depth information by linking depth gradients to RGB values depending on the position within the mesh. 

\begin{figure*}[t!]
\centering
\begin{subfigure}[b]{0.475\textwidth}
    \centering
    \includegraphics[width=\columnwidth]{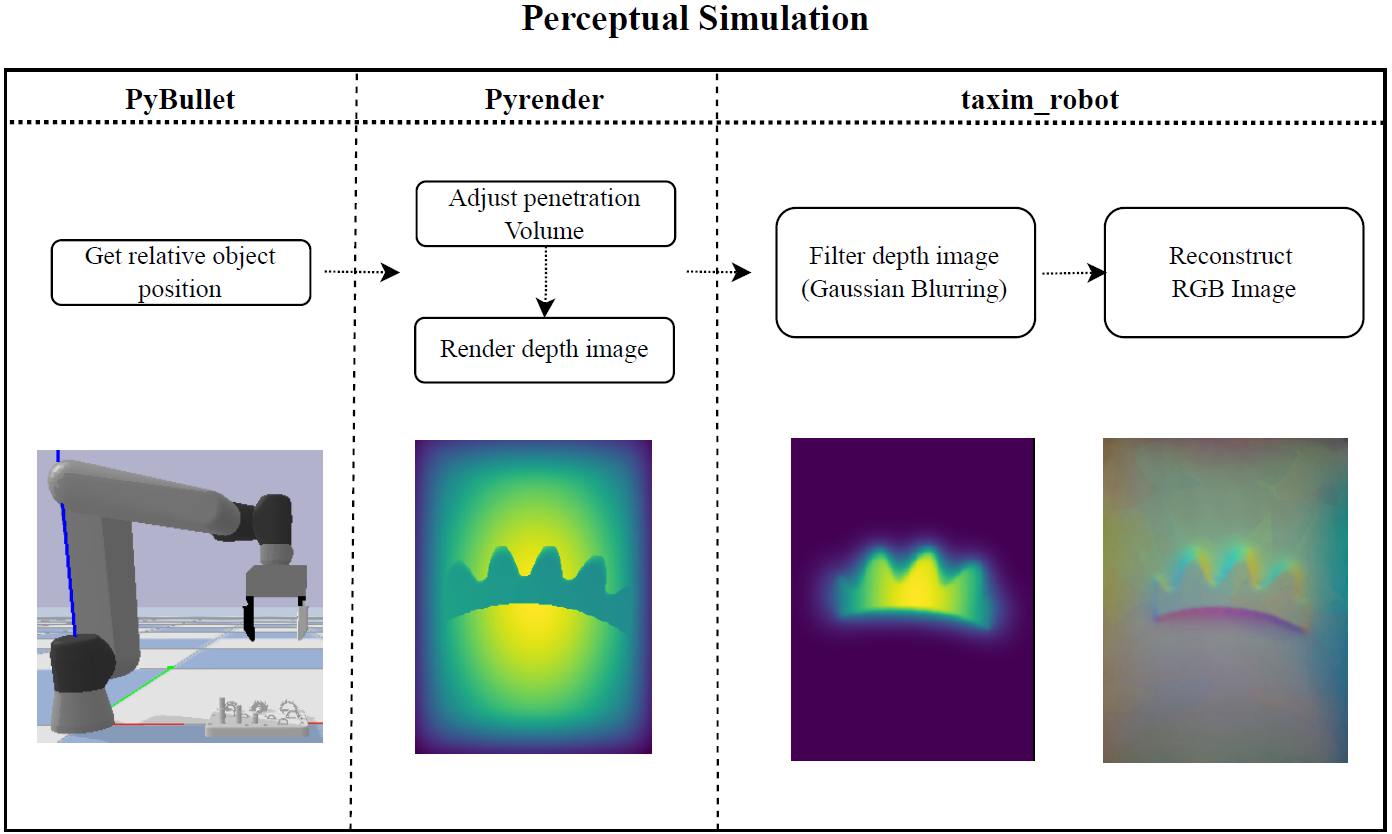}
    \caption{ }\label{fig:optical_sim}
\end{subfigure}
\hfill
\begin{subfigure}[b]{0.475\textwidth}
    \centering
    \includegraphics[width=0.94\columnwidth]{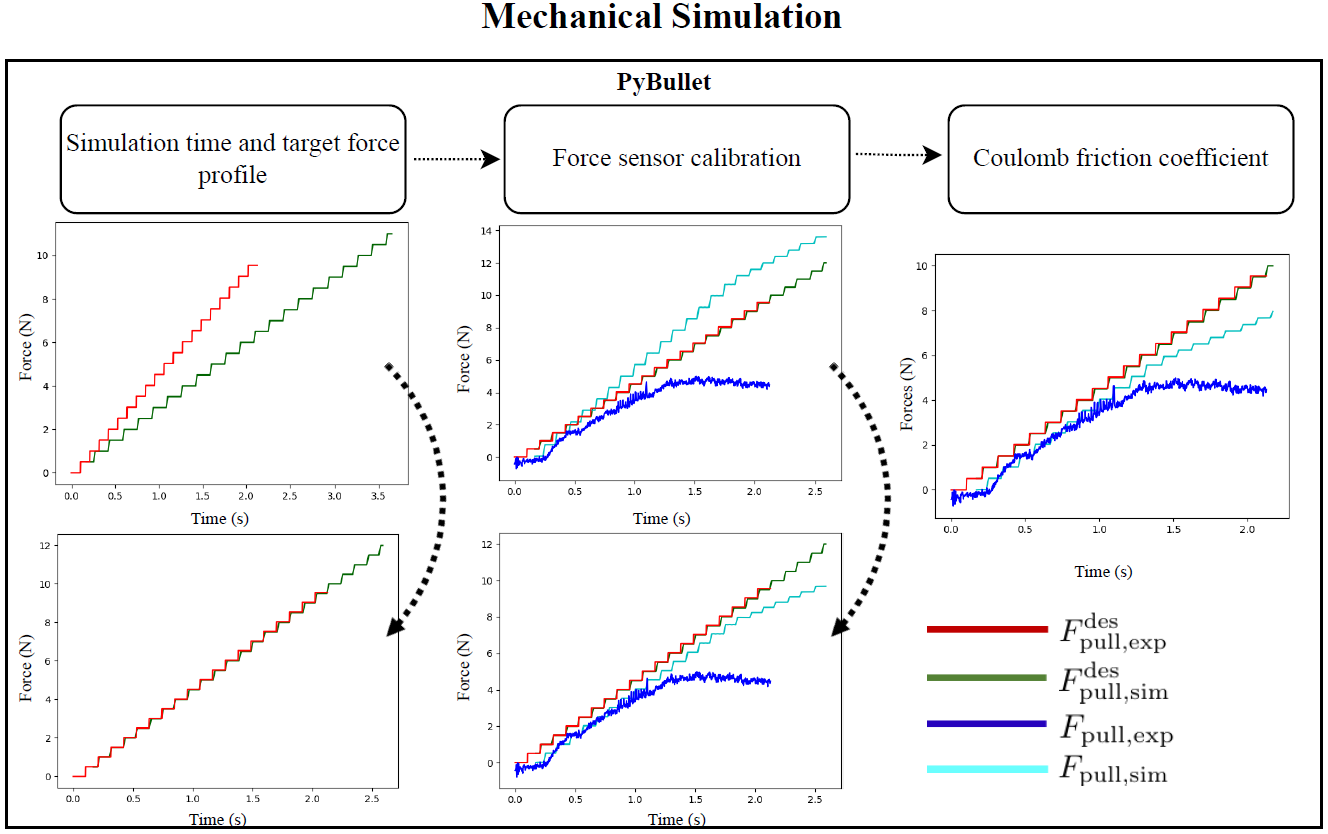}
    \caption{ }\label{fig:mechanical_sim}
\end{subfigure}
\caption{Perceptual and mechanical adaptations of the simulation to the real demonstrator. (a) Perceptual properties: Adjustment of deformation constant in the renderer, Gaussian blurring, and LUT obtained from a calibration procedure using the real sensor. (b) Mechanical properties: Synchronisation of controller, calibration of force measurement with a constant offset over all measurements, and adjustment of friction coefficient.}
\end{figure*}

Next, the process shown in Figure \ref{fig:optical_sim} had to be performed:
\begin{itemize}
    \item \textbf{Step 1: Get relative object position.} The  position of the object with respect to the gel as defined in PyBullet is extracted and transferred to pyRender.
    \item \textbf{Step 2: Adjust penetration volume.}  PyBullet does not allow for intersections of objects. Therefore, taxim\_robot uses the assumption that the volume of the intersection $V$ between the gel mesh and gripping object is proportional to the normal force $F_G$
    \begin{equation}
    V = c\cdot F_G\,.
    \end{equation}
    The constant $c$ is determined by comparing the imprint generated by the object in the experiment with the resulting imprint size in simulation for each object class. 
    \item \textbf{Step 3: Render depth image.} pyRender generates the depth image from the scene. 
    \item \textbf{Step 4: Filter depth image.} Successive Gaussian smoothing with kernel sizes $71$, $51$, $21$, $11$, and $5$, respectively, is applied to model the effect of gel deformation.
    \item \textbf{Step 5: Reconstruct RGB image.} The smoothened depth image is converted into a gradient image. The  RGB image is then calculated using the calibrated LUT. Finally, the background image is added to account for fabrication artefacts of the individual gel mesh.
\end{itemize}

\subsection{Simulation time and target force profile}
The controller in the experiment runs with a non-constant cycle time resulting in non-constant step times for the target force $F_{\rm pull}^{\rm des}$. For this reason, the target force profile for the digital model was set to match the average step time $0.104$~s. Hence, the experimental and simulated target force profiles are almost synchronized, see Figure \ref{fig:mechanical_sim} left column.
\subsection{Mechanical Modelling}\label{sec:mechanical_modelling}
Mechanical simulations were performed using PyBullet as the simulation framework with an OpenAI gym environment setup. First, the mechanical components, i.e. the R1.5 sensor, the WSG-50 gripper and the UR5e robot model, were imported as physical models. The last joint of the UR5e was used as a virtual force sensor available in PyBullet. The objects were fixed by assigning them an infinite mass. CAD-models for gears, shafts and bearings were imported in \texttt{.stl} format into the simulation. The procedure for recording data was identical to that in the experimental setup, i.e. it was implemented as an OpenAI gym policy for the environment. 

After the simulation had been set up, the digital model was tuned to produce results matching the real experiment.
\begin{itemize}
    \item \textbf{Force sensor calibration} The virtual force sensor in the wrist joint of the robot was calibrated since the payload was not accounted for. Comparing measured and simulated data (see the central column of Figure \ref{fig:mechanical_sim}), a correction factor of $0.709$ was identified.
    \item \textbf{Coulomb friction coefficient} PyBullet uses an implicit friction cone model where the friction force $F_F$ is proportional to the normal force $F_G$,
    \begin{equation}
        F_F=\mu_{\rm{Sim}}\cdot F_G \,.
    \end{equation}
 The friction coefficient $\mu_{\rm Sim}$  was determined by running one experiment for each class of objects with a known normal force. The friction coefficient for the object class could thus be determined to match the measured maximum pull force $F_{\rm pull}^{\rm max}$. The resulting friction coefficients for different gripping objects \ref{sec:mechanical_modelling} can be seen in Table \ref{tab:fitresults}.
 \end{itemize}

\section[COMPARISON OF SIMULATED AND REAL DATA]{COMPARISON OF SIMULATED AND MEASURED DATA}\label{sec:comparison}
The simulated data obtained from the digital model (see Section ~\ref{sec:digital-model}) was compared to the sensor output of the experiment for defined grip positions of the robot, grip forces $F_G$, and different gear parts.
\subsection{Tactile Sensor Image Comparison} \label{sec:comparison_sensors}
Three different image comparison metrics,  mean square error (MSE), structured similarity index method (SSIM), and peak signal to noise ratio (PSNR) \cite{wackerly2014mathematical,lu2019level} (see Table \ref{tab:sim_real_ImageCompare}), were computed. The MSE metric can take values within the range $[0,65025]$, PSNR in the range $[0,\infty)$, and SSIM in the interval $[0,1]$. A lower measure for MSE indicates better fit of the images while the opposite holds for PSNR and SSIM. 

The resulting SSIM values of over $0.8$ confirm that the reconstructed RGB images are very similar to the measured images. MSE and PSNR are slightly worse compared to the results presented by Si \textit{et al.}. \cite{si2022taxim} Both metrics are position-dependent which can lead to a bias since the position of the gripper relative to the object is difficult to synchronise exactly. Moreover, the CAD data and the reconstructed geometric data of the gripping objects does not match exactly after the metal parts had been partially reworked for assembly (grinding of the edges) or reconstructed. For example, the teeth of the gear are shaped differently, and the rings inside the ball bearing are of a different size. 

\begin{table}[t!]
\caption{Real sensor images (top) and images simulated using Taxim (bottom) for several parts at different grip strengths. Below each pair are the results of the picture similarity metrics.}
\label{tab:sim_real_ImageCompare}
\centering\begin{tabular}{cccccc} 
 \toprule
 \multicolumn{2}{c}{\textbf{Ball Bearing}}& \multicolumn{2}{c}{\textbf{Gear}}& \multicolumn{2}{c}{\textbf{Long Shaft}}\\
\midrule
$25$~N& $55$~N& $25$~N& $55$~N& $25$~N& $55$~N\\
 \midrule
\includegraphics[width=0.1\linewidth]{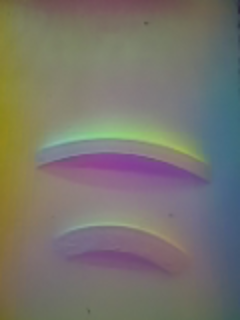}&\includegraphics[width=0.1\linewidth]{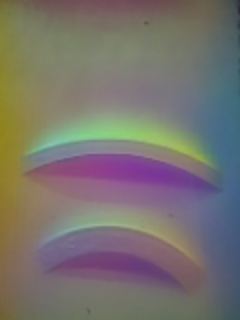}&\includegraphics[width=0.1\linewidth]{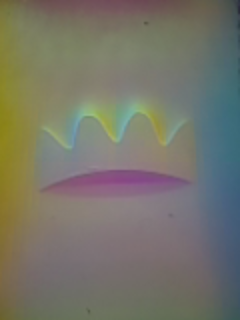}&\includegraphics[width=0.1\linewidth]{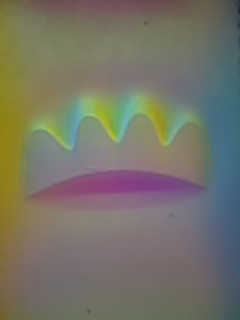}&\includegraphics[width=0.1\linewidth]{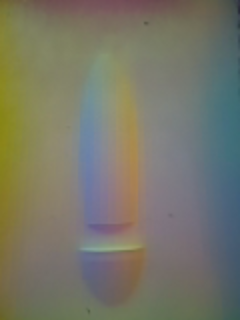}&\includegraphics[width=0.1\linewidth]{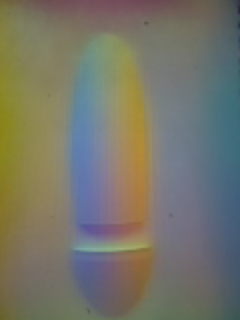}\\
\midrule
\includegraphics[width=0.1\linewidth]{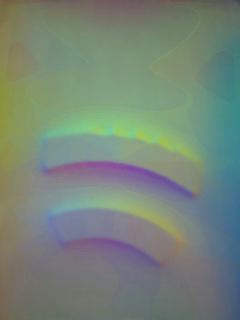}&\includegraphics[width=0.1\linewidth]{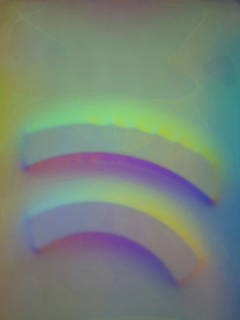}&\includegraphics[width=0.1\linewidth]{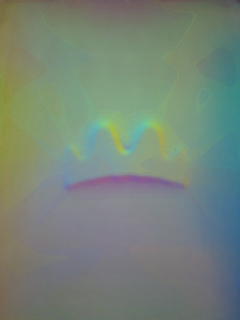}&\includegraphics[width=0.1\linewidth]{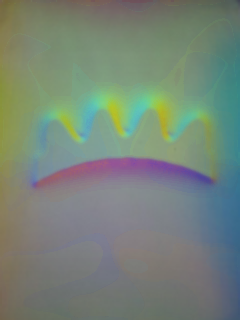}&\includegraphics[width=0.1\linewidth]{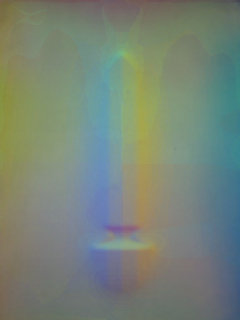}&\includegraphics[width=0.1\linewidth]{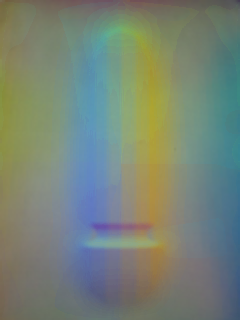}\\
\toprule
\multicolumn{6}{c}{\textbf{MSE}}\\
\midrule
$262.95$&$399.47$&$238.76$&$387.21$&$256.81$&$278.60$\\
\toprule
\multicolumn{6}{c}{\textbf{PSNR}}\\
\midrule
$23.93$&$22.12$&$24.35$&$22.25$&$24.03$&$23.68$\\
\toprule
\multicolumn{6}{c}{\textbf{SSIM}}\\
\midrule
$0.89$&$0.83$&$0.92$&$0.85$&$0.92$&$0.91$\\
\bottomrule
\end{tabular}
\end{table} 

\begin{figure}[t!]
\centering
\begin{subfigure}[b]{0.475\columnwidth}
    \centering
    \includegraphics[width=\columnwidth]{Images/reversedforces_measured.png}
    \caption{Measurement performed in the experiment.}\label{fig:label_real}
\end{subfigure}
\hfill
\begin{subfigure}[b]{0.475\columnwidth}
    \centering
    \includegraphics[width=\columnwidth]{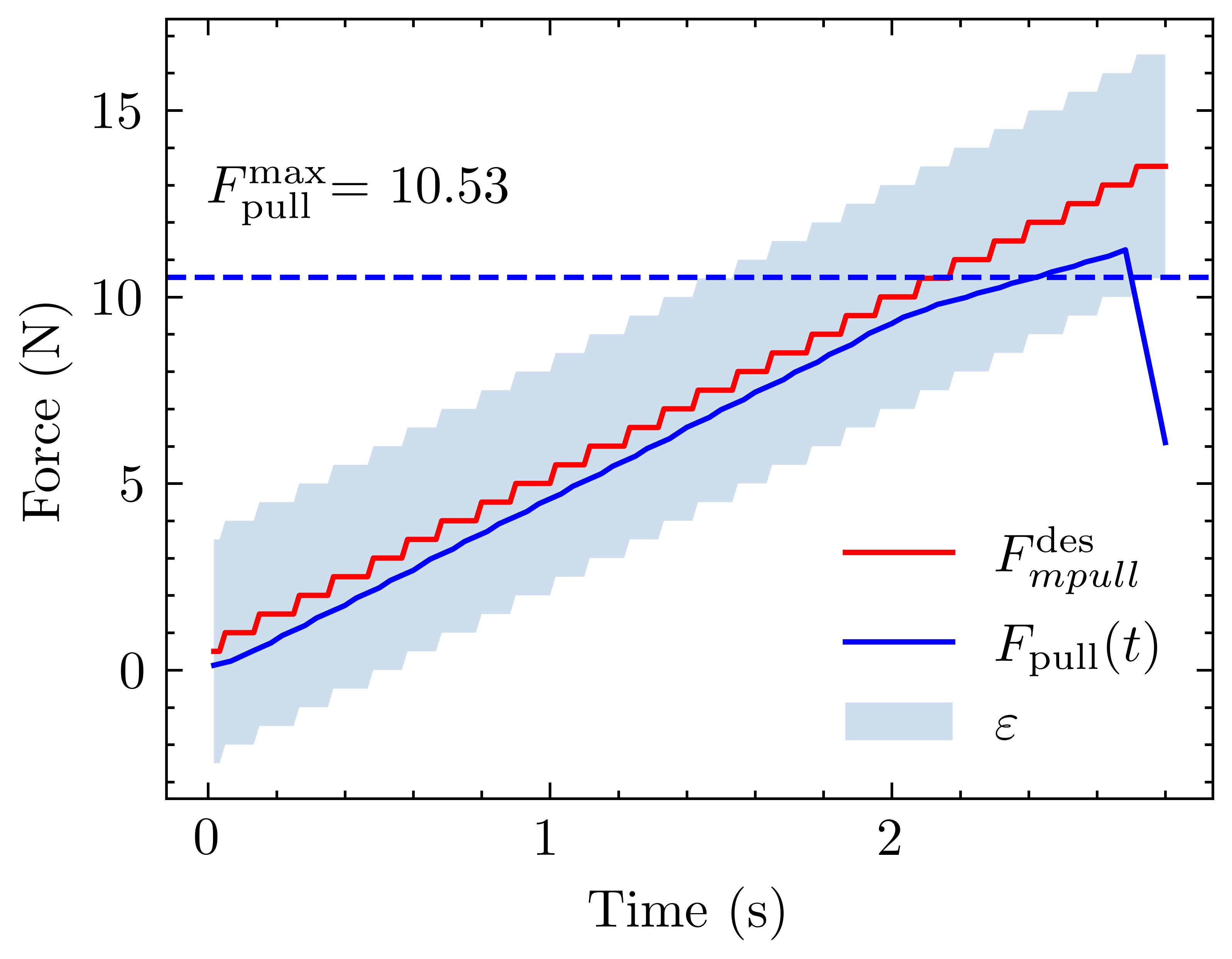}
    \caption{Measurement performed in the simulation}\label{fig:label_sim}
\end{subfigure}
\caption{Comparison between real and simulated pull profiles for $40$~N grip force on a long shaft with lateral friction coefficient of $0.168$.}\label{fig:sim_real_labelCompare}
\end{figure}

\begin{figure*}[t!]
\centering
\begin{subfigure}[b]{0.325\textwidth}
    \centering
    \includegraphics[width=\columnwidth]{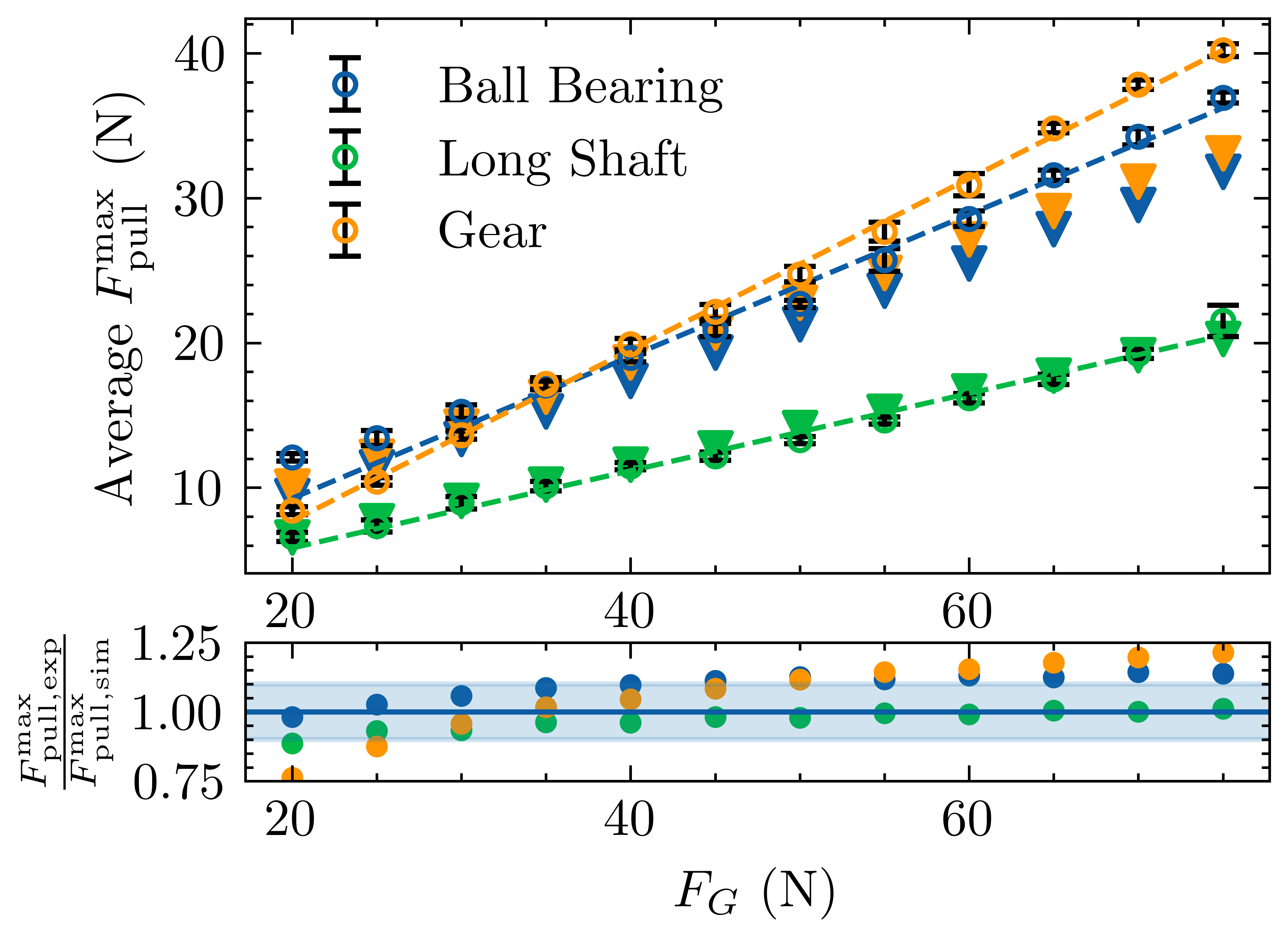}
    \caption{ }\label{fig:comparison_raw}
\end{subfigure}
\hfill
\begin{subfigure}[b]{0.325\textwidth}
    \centering
    \includegraphics[width=0.98\columnwidth]{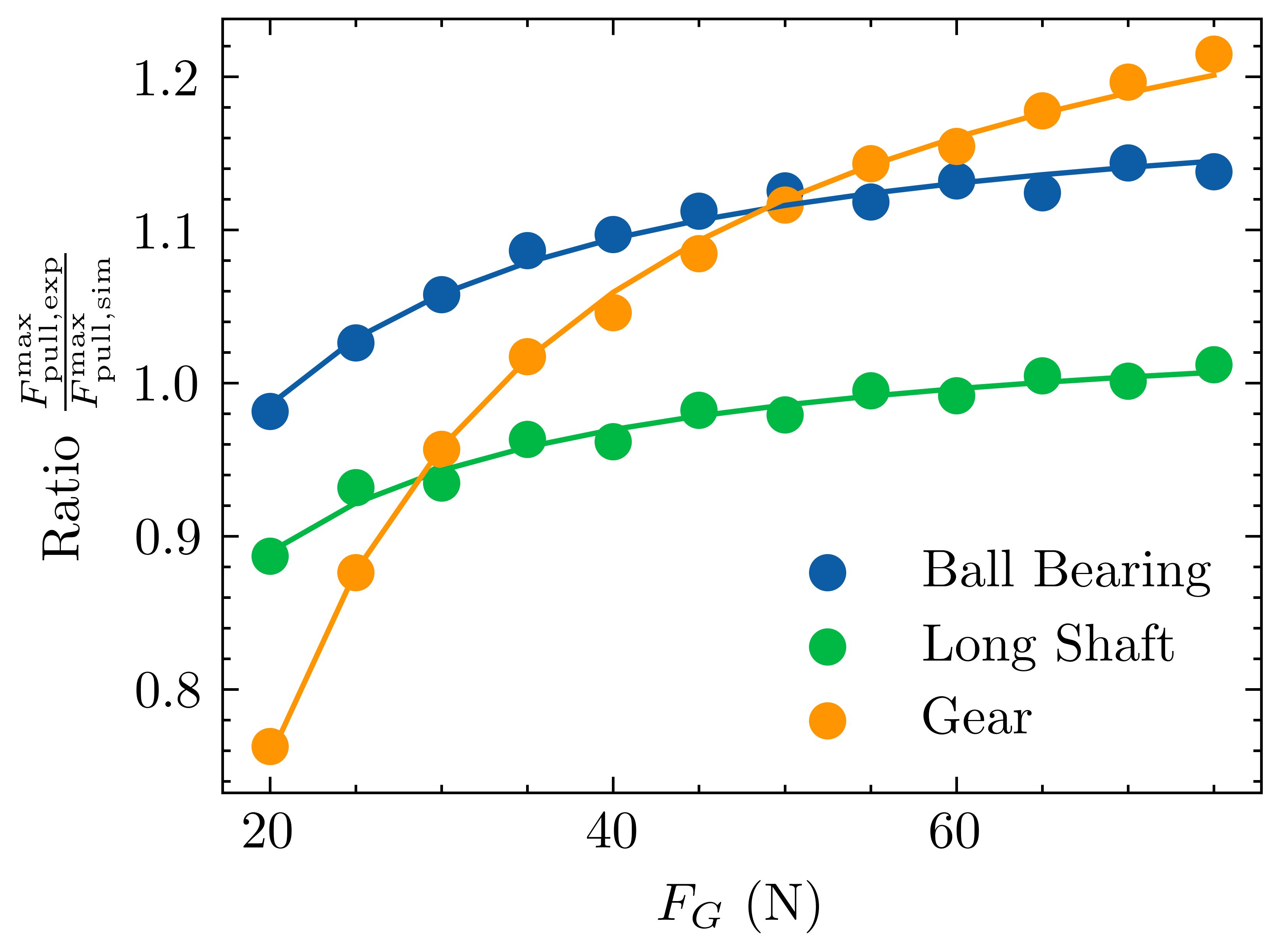}
    \caption{ }\label{fig:ratio}
\end{subfigure}
\hfill
\begin{subfigure}[b]{0.325\textwidth}
    \centering
    \includegraphics[width=\columnwidth]{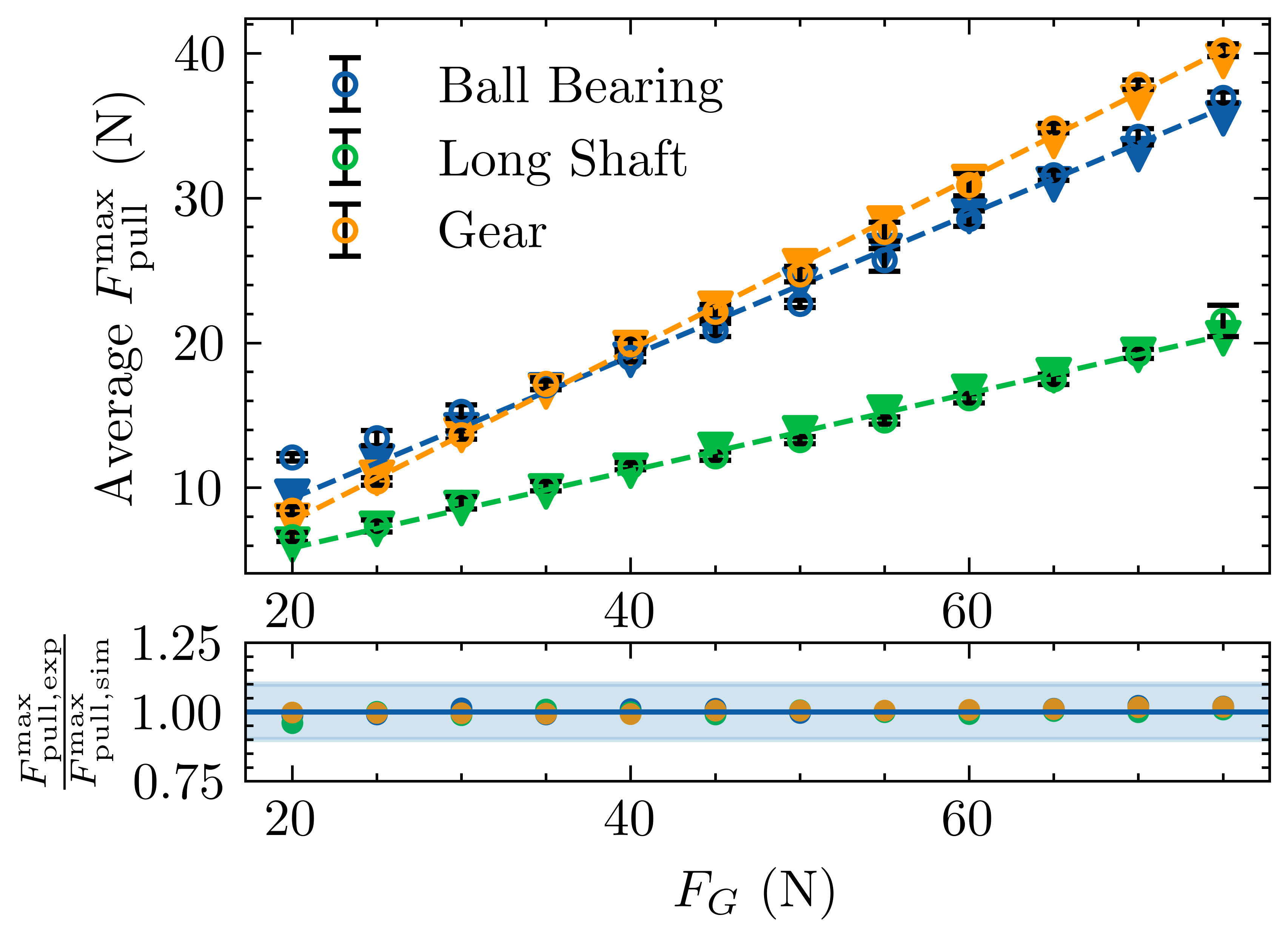}
    \caption{ }\label{fig:comparison_corrected}
\end{subfigure}
\caption{(a) Results of experiment (circles) with linear fit (dashed line) and simulation (triangles). The bottom plot shows the ratio between the linear fit of the experiment and simulation data. (b) The experiment (linear fit) to simulation ratio plot before adjustment. The lines represent the respective fit. (c) Results after the coefficient of friction has been adjusted using the fit parameters obtained in (b).}
\end{figure*}

\subsection{Comparison of Label Calculation}
Figure \ref{fig:sim_real_labelCompare} shows an example of label calculation comparison for a long shaft using a grip force of $40$~N, where both plots exhibit similar behaviour --- The measured forces are linear for lower gripping forces until a certain threshold is reached, indicating the occurrence of slipping, after which the measured forces deviate from the target step profile before slipping off the object completely. Both the simulation and the experiment exhibit different behavior during and after slip. As only the point when the slip occurs is of interest and not what happens afterwards, no mitigation of the effects was required. In the simulation, the robot controller showed an almost instantaneous reaction even for small pull forces close to $0$~N while the UR5 exhibited delayed step responses. Hence, the time lag between set and measured forces is higher in experimental than in simulated data, so that the difference $(F_{\rm pull}^{\rm des} - F_{\rm pull})(t)$ becomes smaller. However, as depicted in Figures \ref{fig:label_real} and \ref{fig:label_sim}, the maximum pull forces are almost identical, $F_{\rm pull}^{\rm max} \approx 11$~N.



\subsection{Comparison of Tensile Profiles at Different Grip Forces}\label{sec:comparison_grip}
 
The Coulomb model does not allow to model the friction phenomena occurring in experiments for all grip configurations. Indeed, a phase transition occurs at around $20$~N in measurements, causing the linear regime above $20$~N not to pass through the origin if extrapolated to low grip forces. This discrepancy between experiment and simulation leads to deviations of  $25\%$ as to the maximum pull force, which can be observed in Figure \ref{fig:comparison_raw}. 
Since PyBullet does not offer more advanced friction models, the only way to mitigate this effect is to adjust the coefficient of friction $\mu_{\rm sim}$ used in the simulation for every grip separately using  
\begin{equation}\label{eq:friction_model_adjusted}
    F_F = \mu'\cdot F_G+F_{F_{\rm off}}\, ,
    \end{equation} 
    where $\mu'$ is a modified coefficient of friction and $F_{F_{\rm off}}$ is a constant friction force offset. 
Using \ref{eq:friction_model_adjusted}, the ratio between the corrected and preset coefficient of friction becomes
\begin{equation}
    \frac{\mu_{\rm corr}}{\mu_{\rm sim}} = \frac{F_{\rm pull,exp}^{\rm max}}{F_{\rm pull,sim}^{\rm max}} = \frac{\mu' F_G+ F_{F_{\rm off}}}{\mu_{\rm sim} F_G} = a + \frac{b}{F_G}. \label{eq:factor}
\end{equation}
 Equation (\ref{eq:factor}) can now be fitted using a ratio plot between the linear fit of the measured and simulated maximum pull forces $F_{\rm pull,exp}^{\rm max}$ and $F_{\rm pull,sim}^{\rm max}$, respectively. The results are shown in Figure \ref{fig:ratio} and Table \ref{tab:fitresults}.  Then, the OpenAI gym environment was adapted to adjust the coefficient of friction $\mu_{\rm corr}$ separately for each grip depending on the grip force set for the simulation $F_G$,
\begin{equation}
\mu_{\rm corr}=\mu_{\rm sim}\cdot \left(a + \frac{b}{F_G}\right)\,
\end{equation}
where $a=\frac{\mu'}{\mu_{\rm sim}}$ and $b=\frac{F_{F_{\rm off}}}{\mu_{\rm sim}}$ as before.
Simulation results using the new coefficients of friction are depicted in Figure \ref{fig:comparison_corrected}, which shows that the difference between the experiment and simulation was reduced to less than $5\%$. 
This is well within the noise level observed in measurements, which justifies to use the synthetic data as training set for a grip stability metric.
\begin{table}[t!]
\caption{Parameters for the correction of coefficients of friction obtained from fitting the ratio of experimental and simulated measurements. }\label{tab:fitresults}
\centering
\begin{tabular}{crrr} 
 \toprule
\textbf{Part} &\multicolumn{1}{c}{$\mu_{\rm sim}$} & \multicolumn{1}{c}{$a$} & \multicolumn{1}{c}{$b$} \\
 \midrule
 Ball Bearing & $0.14$  &$1.20$&$-4.34$  \\

 Long Shaft & $0.168$ &$1.05$& $-3.20$ \\

 Gear & $0.15$ & $1.36$ & $-12.15$ \\ 

 Roller Bearing & $0.14$ & $1.10$ & $10.15$\\
\bottomrule
\end{tabular}
\end{table}

To minimize the data to be taken for each calibration, an investigation on the uncertainty resulting from the number of data points taken has been performed. This can be done by estimating the maximum relative uncertainty of $\mu_{\rm corr}$ using propagation of uncertainty and the covariance matrix of the fit result. With a minimum of three data points collected in the real experiment for tuning the simulation, a maximum relative error of $\frac{\sigma_{\mu_{corr}}}{\mu_{corr}}=3\%$ occurs. In the case of the full data set, the relative error is $\frac{\sigma_{\mu_{corr}}}{\mu_{corr}}=1.1\%$. In both cases it is below the 5\% significance level, so it should be sufficient to take 3 data points with the experiment to calibrate the simulation for an object. Since the contact properties are slightly different for each part, the adjustment must be performed for each object geometry. 

\section{Conclusions and Outlook}\label{sec:outlook}
Generating a diverse data set to train a neural grip metric network using experiments alone is time consuming. The digital model implemented in this work automatically generates accurate synthetic training data. Tuning the free parameters in the simulation took considerable effort and produced results that represents the actual experiment as closely as possible. The optical simulation of the tactile sensor using an adapted version of Taxim is in good agreement as the similarity metric SSIM proved. As to the mechanical modeling, the simplified friction model available in the simulator requires at least three data points per object
to reproduce maximum pull forces correctly. As a result, the simulated forces above $20$~N matched the experimental data up to measurement noise. Further investigations will be performed to understand the non-linearities observed in the grip force regime below $20$~N to be able to extend the simulated data sets. 
In order to further reduce the amount of real data to optimize the mechanical properties of the simulation, it will be investigated whether the geometric properties of the objects can be classified. 

The simulation presented in this paper will play a pivotal role to develop an advanced grip stability metric in the future. The approach will train a neural network  using the RGB images obtained from the tactile sensor before the object has been lifted labeled with the measured pull forces $F_{\rm pull}^{\rm max}$. 
Both the experimental data and the synthetic data will be trained separately with the different instances of the same neural network. The achieved models will be compared with each other by evaluating their training performance and prediction power. Once the synthetic data set has been validated, the data sets are to be combined and extended with a wide range of parts used in industrial applications including synthetic and experimental data to train a hybrid model.
The digital twin presented in this paper provides the framework to achieve this.  

\bibliographystyle{unsrt}
\bibliography{Wucherer_arXiv}

\end{document}